\documentclass[11pt]{article}
\usepackage{coling2020}
\usepackage{times}
\usepackage{url}
\usepackage{latexsym}
\usepackage{graphicx}
\usepackage{caption}
\usepackage{subcaption}
\usepackage{tabularx}
\usepackage{xstring}
\usepackage[textsize=tiny]{todonotes}
\usepackage{booktabs}
\usepackage{amsmath}
\usepackage[T1]{fontenc}
\usepackage{fancyvrb}
\usepackage{bera}
\usepackage{array}
\usepackage{makecell}
\usepackage{rotating}
\usepackage{tikz}
\usepackage{hyperref}
\usepackage[utf8]{inputenc}

\usepackage{fancyhdr}
\usepackage{lipsum}% just to generate text for the example

\fancypagestyle{specialfooter}{%
  \fancyhf{}
  
  \fancyfoot[C]{\small In Proceedings of the International Digital Humanities Conference DH2020, Ottawa. \\ \url{https://dh2020.adho.org/wp-content/uploads/2020/07/600_MappingTopicEvolutionAcrossPoeticTraditions.html}}
}

\title{Mapping Topic Evolution Across Poetic Traditions}

\author{
  Petr Plechac \\[3pt]
  Institute of Czech Literature, \\
  Czech Academy of Sciences, Prague \\[3pt]
  University of Basel \\[3pt]
  {\tt plechac@ucl.cas.cz}
  {\tt } \\\And
  Thomas Haider \\[3pt]
  Max Planck Institute for \\ Empirical Aesthetics, Frankfurt \\[3pt]
  University of Stuttgart \\[3pt]
  {\tt thomas.haider@ae.mpg.de}\\
   \\}

\date{}

\begin{document}
\thispagestyle{specialfooter}
\maketitle
\begin{abstract}
Poetic traditions across languages evolved differently, but we find that certain semantic topics occur in several of them, albeit sometimes with temporal delay, or with diverging trajectories over time. We apply Latent Dirichlet Allocation (LDA) to poetry corpora of four languages, i.e., German (74k poems), English (85k poems), Russian (18k poems), and Czech (80k poems). We align and interpret salient topics, their trend over time (1600--1925 A.D.), showing similarities and disparities across poetic traditions with a few select topics, and use their trajectories over time to pinpoint specific literary epochs with a focus on Romanticism and Modernity.
\end{abstract}

\section{Corpora \& Model}

To determine the evolution of topics across poetic traditions, we collect four poetry corpora in Czech, Russian, German and English. See Table \ref{tab:corpora} for an overview, and where they were mined from. As these corpora are often contaminated with foreign language poems, we filter these with {\tt{langdetect}}.\footnote{\url{https://pypi.org/project/langdetect/}}

\begin{table}[ht]
 \center 
\begin{tabular}{ r c c l }
\toprule
  Language        &  Poems  &  Tokens & Comment \\
\toprule
      Czech &     ${\sim}80k$   &   15M &  \makecell[l]{Corpus of Czech Verse \\ ({\url{http://versologie.cz}})} \\[10pt]
      
      Russian & ${\sim}18k$   &   2.7M  & \makecell[l]{Poetic subcorpus of Russian National Corpus \\ ({\url{http://ruscorpora.ru}})}\\[10pt]
    
      German &    ${\sim}74k$     &   12M & \makecell[l]{German Poetry Corpus v3, Textgrid + Deutsches Textarchiv  \\ ({\url{http://github.com/tnhaider/DLK}})}\\[10pt]
    
      English &    ${\sim}85k$    &   22M &  \makecell[l]{Project Gutenberg, Mined with GutenTag 'Poetry' \\ (\url{https://gutentag.sdsu.edu/})} \\
       \bottomrule
\end{tabular}
\caption{Diachronic Poetry Corpora}
    \label{tab:corpora}
\end{table}

%To filter out poems written in languages foreign to a given corpus, we've employed \href{https://pypi.org/project/langdetect/}{[langdetect]} python library, which is a port of Google's \href{https://code.google.com/archive/p/language-detection/}{[language-detection]}\footnote{\url{https://code.google.com/archive/p/language-detection/}}. 

To learn semantic topics, Latent Dirichlet Allocation (LDA) \cite{blei2003latent} has proved useful. We use the vanilla LDAMultiCore implementation as it is provided in {\tt{genism}}\footnote{\url{https://radimrehurek.com/gensim/models/ldamulticore.html}} \cite{vrehuuvrek2011gensim}. LDA assumes that a particular document contains a mixture of few salient topics of semantically related words. 

We transform our documents  %(of lemmas)
to a bag of words representation\footnote{As we deal also with highly inflected languages (Czech, Russian), lemmas were used instead of word forms. For lemmatization and POS-tagging of English and German texts we use the \href{https://www.cis.uni-muenchen.de/~schmid/tools/TreeTagger/}{\tt{TreeTagger}} \cite{schmid1994treetagger}, for lemmatization and POS-tagging of Czech texts we use the \href{https://ufal.mff.cuni.cz/morphodita}{\tt{MorphoDita}} \cite{strakova2014morphodita}, for lemmatization of Russian texts we use the \href{https://github.com/nlpub/pymystem3}{\tt{MyStem}} \cite{Segalovich2003AFM}. In Czech, German, and English all the parts-of-speech except for nouns, adjectives, and verbs were filtered out.%\todo{Why? Just to get rid of meaningless rubbish}
In Russian, the list of stopwords is provided by the \href{https://www.nltk.org/}{\tt{NLTK}} library and manually extended by us.}  and set the desired number of topics=100 and train for 100 epochs (passes) to attain a reasonable distinctness of topics. We choose 100 topics % (rather than a lower number that might be more straightforward to interpret) 
as previous research on poetic topics \cite{haider2019diachronic,navarro2018poetic} determined this parameter to be be optimal for distant reading.

\section{Experiment Setup}
We approach diachronic variation in poetry as distant reading task to visualize the development of interpretable topics over time and across languages. We retrieve the most important (likely) words for all topics and interpret these (sorted) word lists as aggregated topics. We are then able to manually translate several topics that align over all four corpora.

\begin{figure}[!htb]
    \centering
    \begin{minipage}[t]{.5\textwidth}
        \centering
      \includegraphics[width=1.0\textwidth]{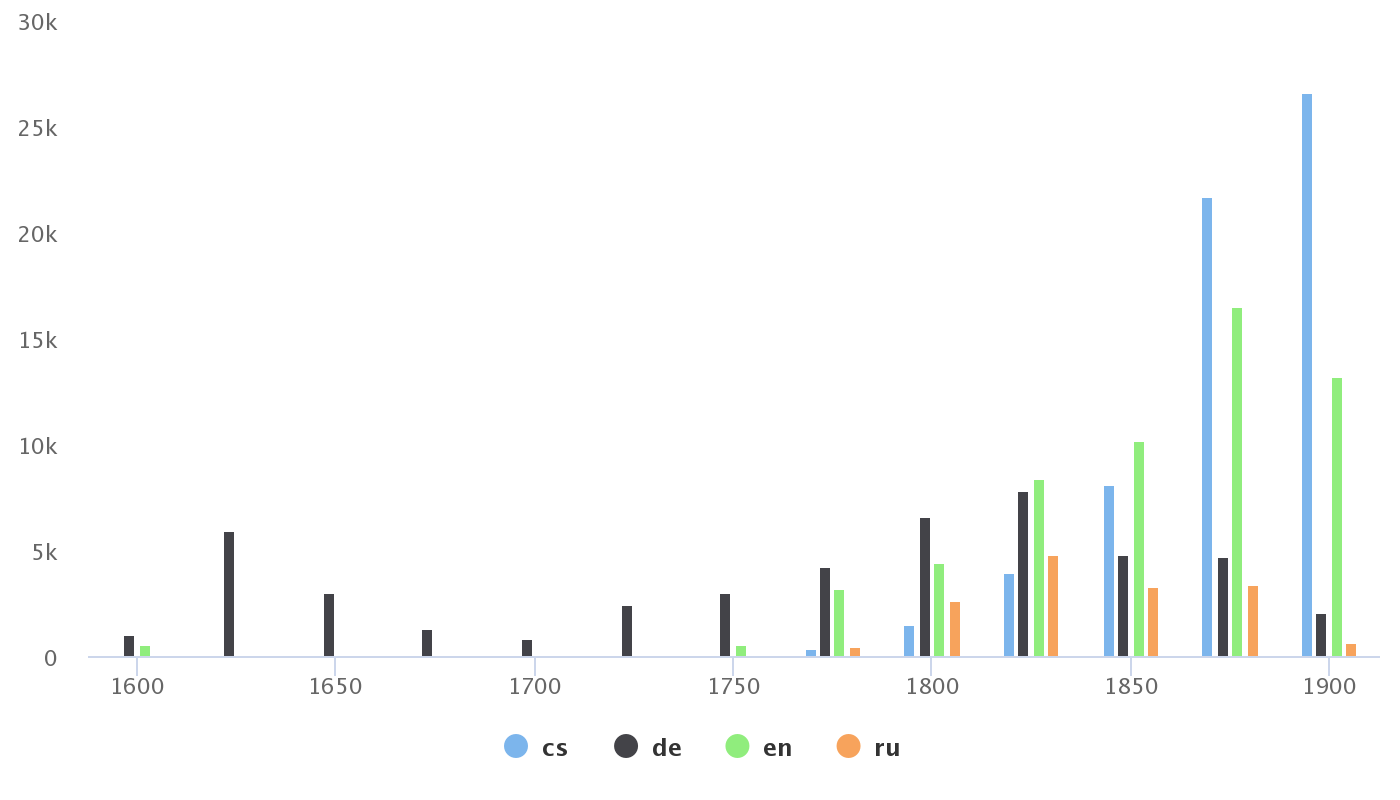}
  \caption{Size of Corpora over Time}
        \label{fig:corpussize}
    \end{minipage}%
    \begin{minipage}[t]{0.5\textwidth}
        \centering
      \includegraphics[width=1.0\textwidth]{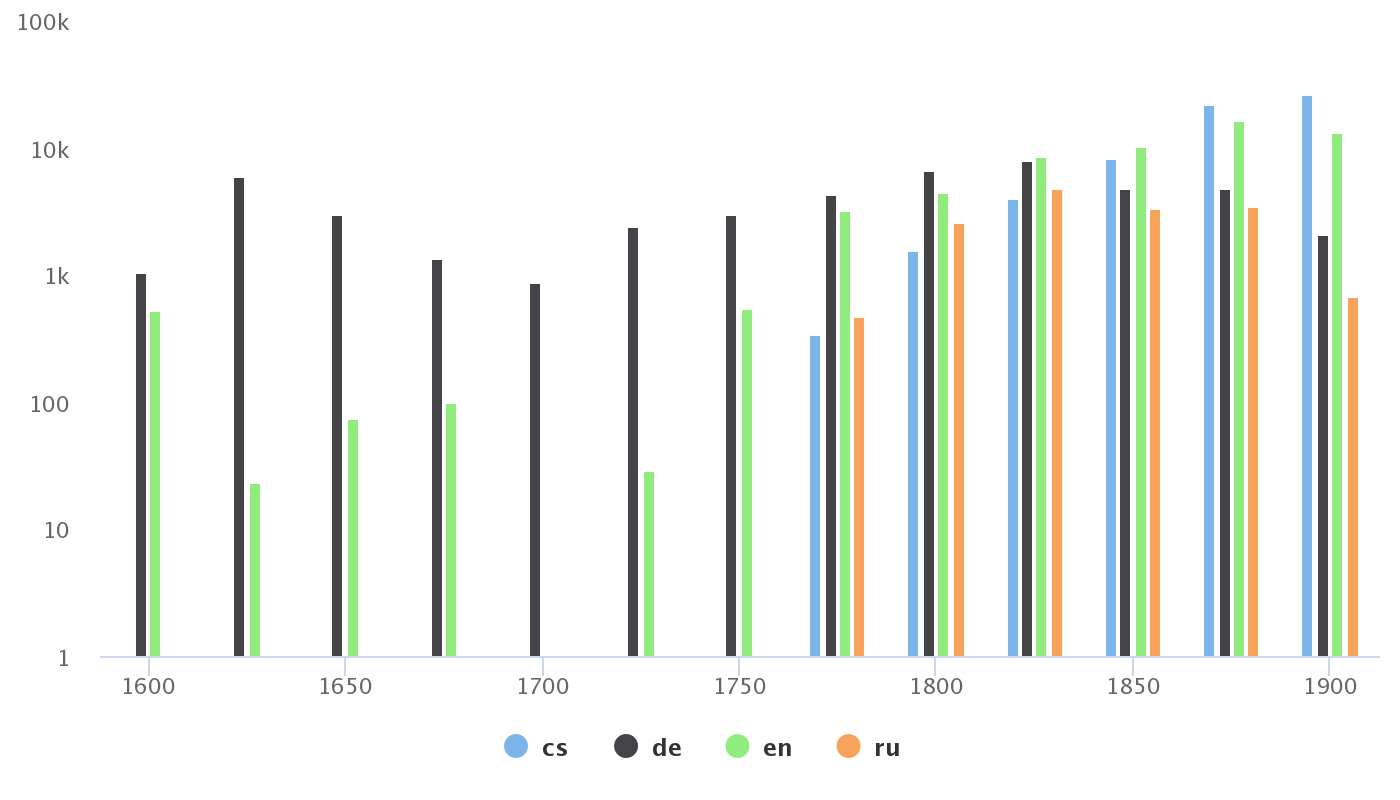}
  \caption{Size of Corpora over Time; log(Size) at y-axis}
        \label{fig:logcorpussize}
    \end{minipage}
\end{figure}

To discover trends over time, we bin our documents into time slots of 25 years width each, except for early English where two large slots (1600–1674 and 1675–1749) were used due to sparse data. See Figures \ref{fig:corpussize} and \ref{fig:logcorpussize} for a plot of the number of documents per bin. %The chosen binning slots offer enough documents per slot for our experiments.
%\todo{what happened before 1700? There is no size statistic. Is is too few documents that the topic trajectories become noisy?} 
To visualize trends of singular topics over time, we follow the strategy of \newcite{haider2019diachronic}: We aggregate all documents \texttt{d} in slot \texttt{s} and sum the probabilities of topic \texttt{t} given \texttt{d} and divide by the number of all \texttt{d} in \texttt{s}. This gives us the average probability of a topic per time slot. We then plot the trajectories for each single topic.

\section{Alignment and Interpretation of topic Trajectories}

\begin{figure}[!htb]
  \centering
      \includegraphics[width=0.8\textwidth]{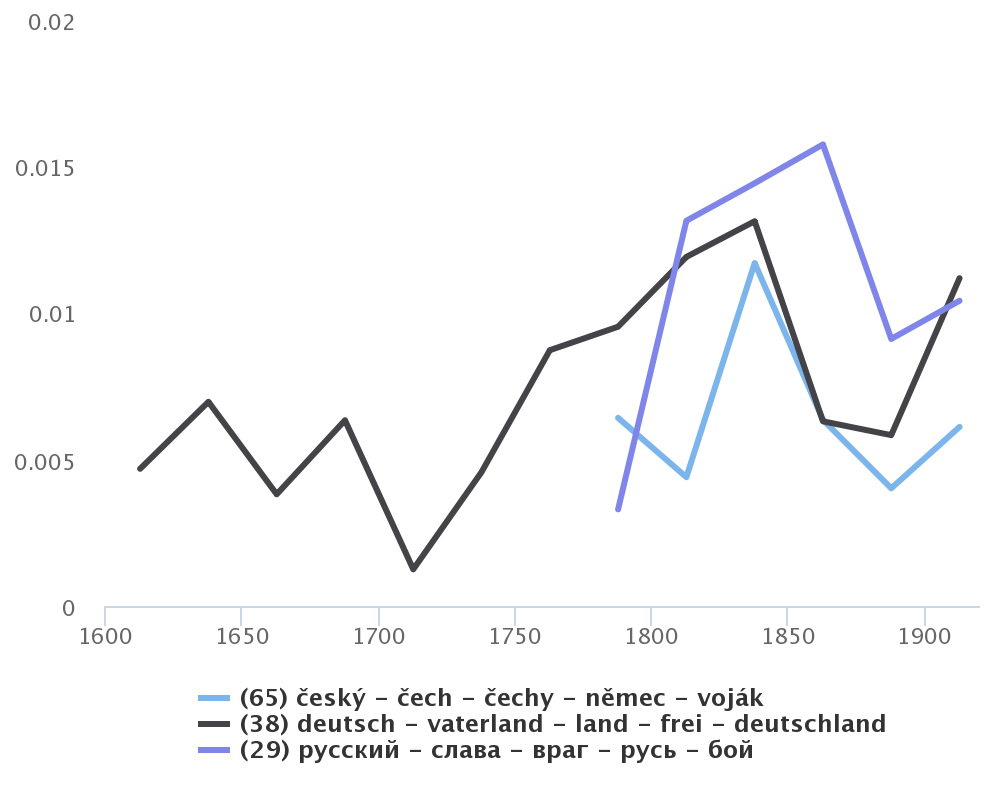}
  \caption{Topic \textbf{Nation}}
  \label{fig:nation}
\end{figure}

\begin{figure}[!htb]
  \centering
      \includegraphics[width=0.8\textwidth]{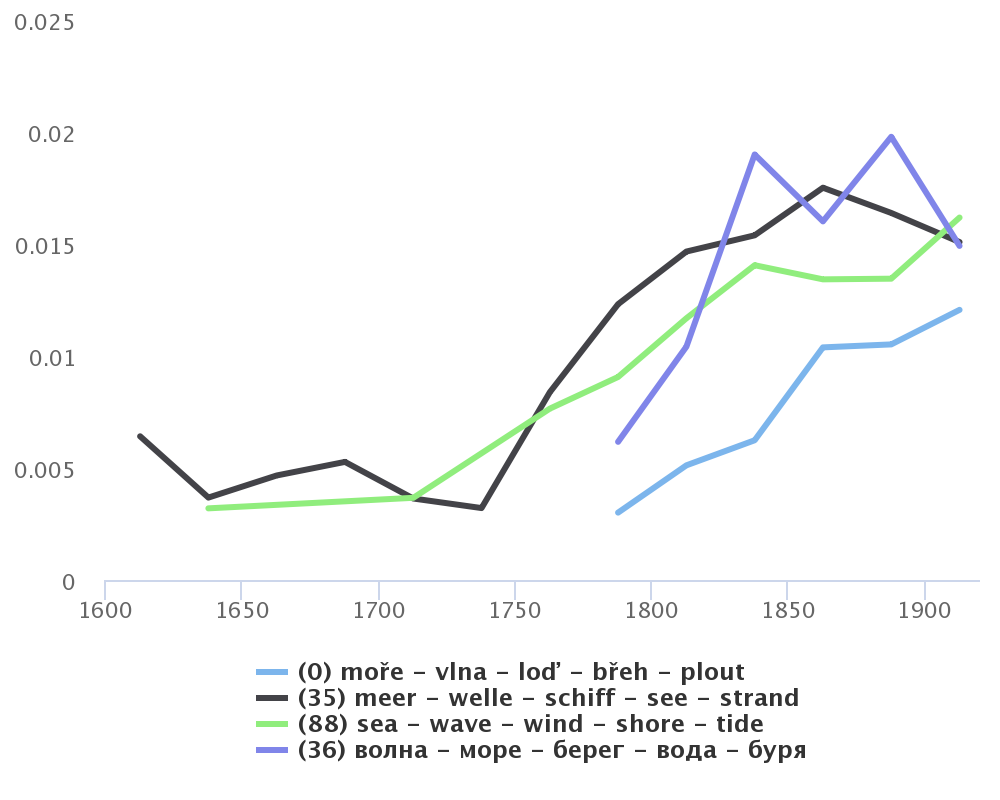}
  \caption{Topic \textbf{Sea}}
  \label{fig:sea}
\end{figure}

\begin{figure}[!htb]
  \centering
      \includegraphics[width=0.7\textwidth]{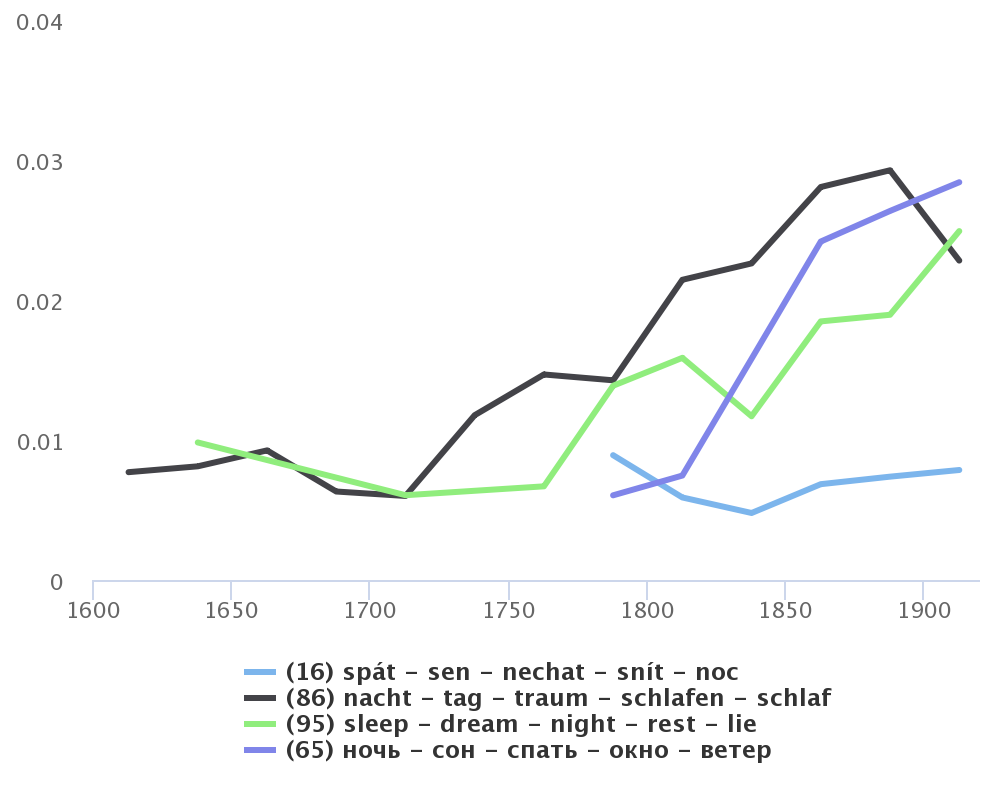}
  \caption{Topic \textbf{Sleep}}
  \label{fig:sleep}
\end{figure}

\begin{figure}[!htb]
  \centering
      \includegraphics[width=0.7\textwidth]{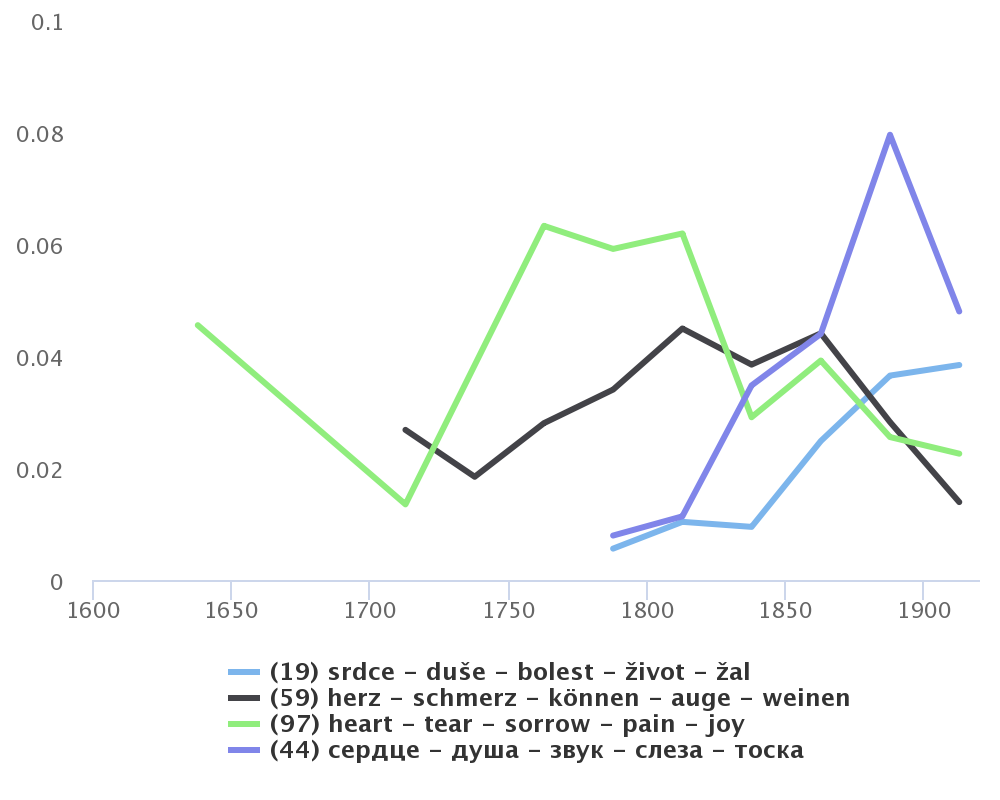}
  \caption{Topic \textbf{Sorrow}}
  \label{fig:sorrow}
\end{figure}

\begin{figure}[!htb]
  \centering
      \includegraphics[width=0.65\textwidth]{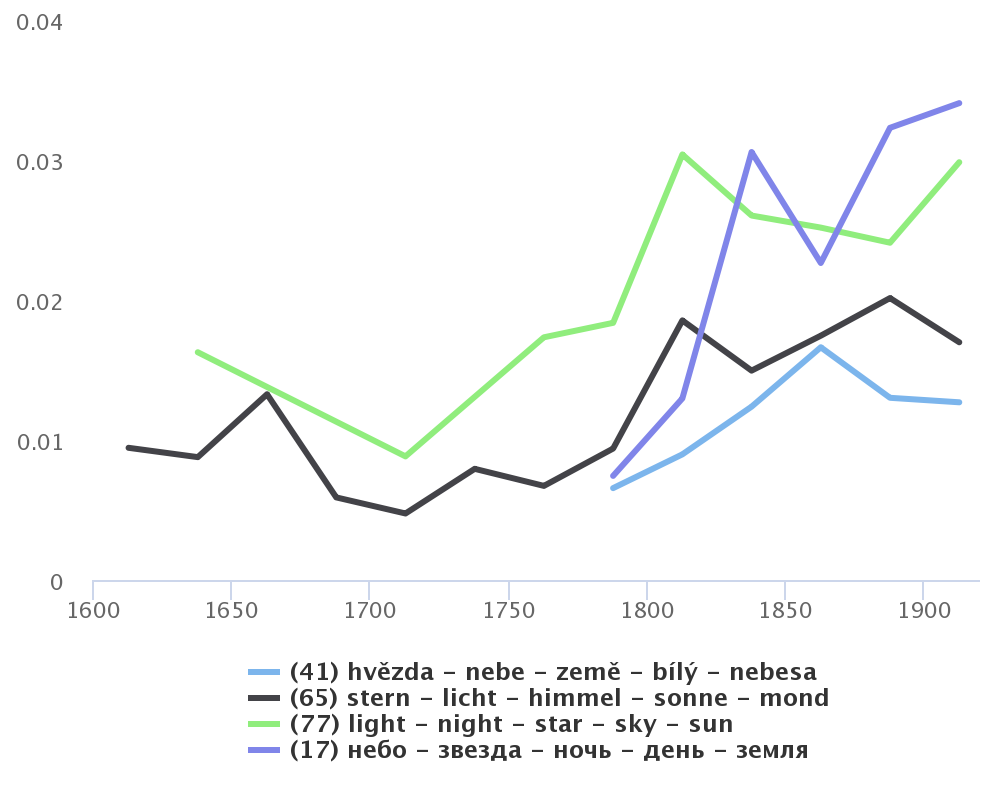}
  \caption{Topic \textbf{Stars}}
  \label{fig:star}
\end{figure}

\begin{figure}[!htb]
  \centering
      \includegraphics[width=0.65\textwidth]{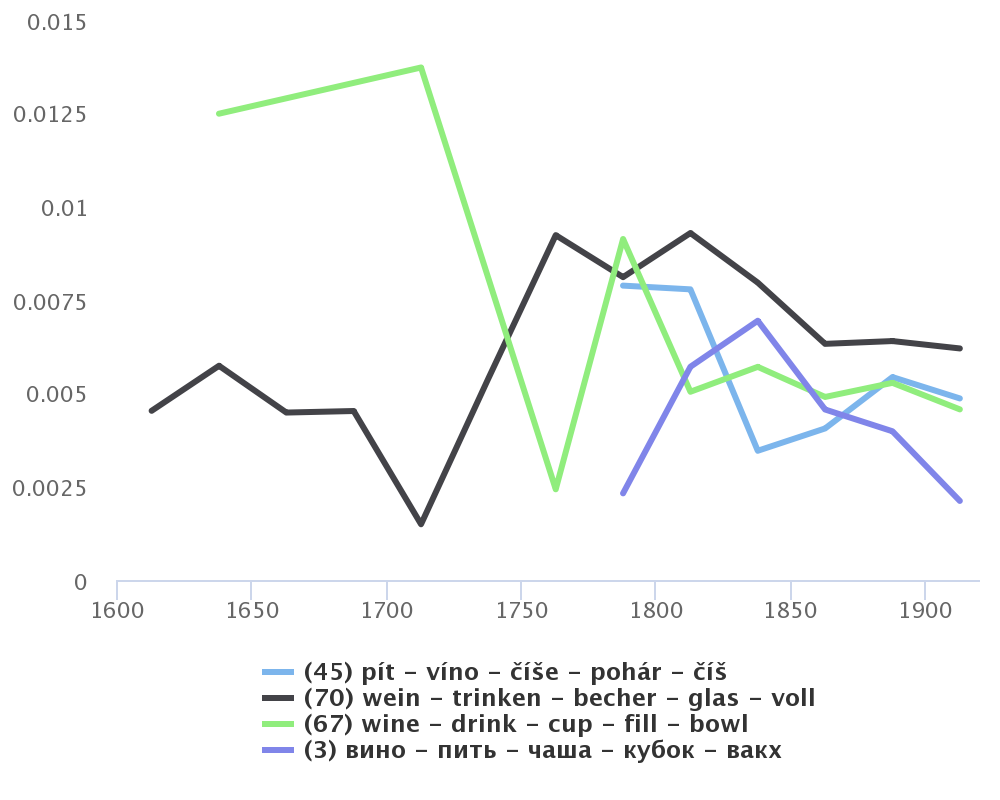}
  \caption{Topic \textbf{Wine}}
  \label{fig:wine}
\end{figure}

Based on a few selected topics, we can trace similarities and disparities over poetic traditions. See Figures \ref{fig:nation}--\ref{fig:wine} for a selection of interpretable topic trends, where the four languages align. %Please note that the scaling on the y-axis differ for each topic, as some topics are more pronounced in the whole dataset overall.

Figure \ref{fig:nation} shows the topic "Nation", which has a similar trend in German, Czech, and Russian, but is not present in the English corpus (cf. completely different geopolitical situation of the British empire). In the German corpus it emerges in the second half of the 18th century and peaks around 1825 to 1850 (outlining the period of `Vormärz'). The same peak can be found in the Czech corpus (late National Revival), and slightly delayed in Russian. In all the three corpora, it loses importance after 1850/60, but is gaining traction once again at the beginning of the 20th century.

Figure \ref{fig:sea} shows the topic "Sea", which has a similar rising tendency towards the second half of the 19th century and stays stable into Modernity. This topic is most pronounced for the Russian and German period of Romanticism, after which it seems to taper off, while it still shows an upward trajectory for English and Czech.

%This includes e.g. the ``sea topic'':

%\begin{itemize}
%    \item moře, vlna, loď, břeh, plout
%    \item meer, land, sturm, schiff, welle
%    \item sea, wave, wind, shore, tide
%    \item \begin{otherlanguage*}{russian}волна, море, берег, вода, буря\end{otherlanguage*} 
%\end{itemize}

\clearpage

The topic "Sleep" (Figure \ref{fig:sleep}) appears rather correlated with the topic "Sea" in English, German, and Russian, with a focus on late Romanticism and Modernity, but it is rather marginal in the Czech corpus.

Figure \ref{fig:sorrow} shows the topic "Sorrow" that has clearly separable trends, with English and German on one hand and Czech and Russian on the other. In the first case it is associated with the period of Romanticism (although becoming prominent earlier in English), and in the latter with late 19th century Modernism (although in Russian it emerges already in the period of Romanticism; 1825 to 1850).

Figure \ref{fig:star} shows the topic "Stars", pronounced in English and German High Romanticism (1800 to 1825) and in Russian Late Romanticism (1825 to 1850). In Czech the peak occurs delayed in the generation of "Máj" (period 1850 to 1875). Note, that these authors claim themselves as the followers of Karel Hynek Mácha (1810–1836), who in turn is well-known for bringing English Romanticism themes into Czech poetry.

Lastly, Figure \ref{fig:wine} shows the topic "Wine" which is clearly associated with the Anacreontics. It is accented in early 18th century English poetry, second half 18th century German poetry, and late 18th century Czech poetry (almanacs edited by A. J. Puchmajer). In Russian poetry it surprisingly peaks in the period of romanticism (1825 to 1850).

%or the  ``wine-drinking topic'':

%\begin{itemize}
%    \item pít, víno, číše, pohár, číš
%    \item wein, trinken, becher, glas, voll
%    \item wine, drink, cup, fill, bowl
%    \item \begin{otherlanguage*}{russian}вино, пить, чаша, кубок, вакх\end{otherlanguage*} 
%\end{itemize}

\section{Conclusion \& Future Work}
In this paper we used Latent Dirichlet Allocation for a visualization of topic trends across languages, illustrating the similarities and disparities between different poetic traditions. Our method is largely based on reading and translating topic distributions and finally interpreting the trajectories of relative topic importance against the backdrop of literary history. We find that some topics, especially the examples chosen, do align across languages, sometimes with temporal delay (as they were picked up later in another language), while other topics were not as heavily discussed in other poetic discourses (such as "Nation" in English). Future work should look into cross-lingual alignment methods, e.g., through multi-lingual embeddings or poly-lingual topic models without parallel data. Finally, over- or underrepresentation of certain authors or near-duplicates of poems (different editions) can lead to corpus imbalance. Consequently, this impacts our measure to calculate the relative importance of a topic given a certain time stamp and should be addressed in future work.

%\section{Bibliographical References}
\label{main:ref}

\bibliographystyle{acl}
\bibliography{biblio}

\begin{thebibliography}{}

\bibitem[\protect\citename{Blei \bgroup et al.\egroup }2003]{blei2003latent}
David~M Blei, Andrew~Y Ng, and Michael~I Jordan.
\newblock 2003.
\newblock Latent dirichlet allocation.
\newblock {\em Journal of machine Learning research}, 3(Jan):993--1022.

\bibitem[\protect\citename{Haider}2019]{haider2019diachronic}
Thomas~N Haider.
\newblock 2019.
\newblock Diachronic topics in new high german poetry.
\newblock {\em Proceedings of the International Digital Humantities Conference
  DH2020 in Utrecht}.

\bibitem[\protect\citename{Navarro-Colorado}2018]{navarro2018poetic}
Borja Navarro-Colorado.
\newblock 2018.
\newblock On poetic topic modeling: extracting themes and motifs from a corpus
  of spanish poetry.
\newblock {\em Frontiers in Digital Humanities}, 5:15.

\bibitem[\protect\citename{Rehurek and Sojka}2011]{vrehuuvrek2011gensim}
Radim Rehurek and Petr Sojka.
\newblock 2011.
\newblock Gensim—statistical semantics in python.
\newblock {\em statistical semantics; gensim; Python; LDA; SVD}.

\bibitem[\protect\citename{Schmid}1994]{schmid1994treetagger}
Helmut Schmid.
\newblock 1994.
\newblock Probabilistic part-of-speech tagging using decision trees.
\newblock In {\em Proceedings of International Conference on New Methods in
  Language Processing}, Manchester, UK.

\bibitem[\protect\citename{Segalovich}2003]{Segalovich2003AFM}
Ilya Segalovich.
\newblock 2003.
\newblock A fast morphological algorithm with unknown word guessing induced by
  a dictionary for a web search engine.
\newblock In {\em MLMTA}.

\bibitem[\protect\citename{Straková \bgroup et al.\egroup
  }2014]{strakova2014morphodita}
Jana Straková, Milan Straka, and Jan Hajič.
\newblock 2014.
\newblock Open-source tools for morphology, lemmatization, pos tagging and
  named entity recognition.
\newblock In {\em Proceedings of 52nd Annual Meeting of the Association for
  Computational Linguistics: System Demonstrations}, pages 13--18, Baltimore,
  Maryland, jun. Association for Computational Linguistics.

\end{thebibliography}

\end{document}